\documentclass[superscriptaddress,floatfix]{article}

\usepackage{etex} 

\usepackage[all]{xy}
\usepackage{xargs}
\usepackage{listings}
\usepackage[utf8]{inputenc}
\usepackage[text={7in,8.0in},centering]{geometry}
\usepackage{amsmath}
\usepackage{amssymb}
\usepackage{amsthm}
\usepackage[ampersand]{easylist}
\usepackage{bbm}
\usepackage{tabularx}
\usepackage{url}
\usepackage{fancyhdr}
\usepackage{algpseudocode}
\usepackage{algorithm}
\usepackage{empheq}
\usepackage{color}
\usepackage{graphicx}
\usepackage{hyperref}
\usepackage{caption}
\usepackage{subcaption}
\usepackage{tikz}
\usepackage[parfill]{parskip} 
\usepackage{scalerel}
\usepackage{etoolbox}

\definecolor{lightyellow}{cmyk}{0,0,0.3,0}
\graphicspath{{../Figures/}{./Figures/}{./}}

\newcommand{\fwd}{\textrm{fwd}}
\newcommand{\bck}{\textrm{bck}}

\newcommand{\SO}{\textrm{SO}}
\newcommand{\U}{\textrm{U}}

\newcommand{\vc}[1]{{\pmb{#1}}}

\newcommand{\Av}{\vc{A}}

\newcommand{\cv}{\vc{c}}

\newcommand{\gv}{\vc{g}}
\newcommand{\Gv}{\vc{G}}

\newcommand{\Iv}{\vc{I}}

\newcommand{\Mv}{\vc{M}}

\newcommand{\Pv}{\vc{P}}
\newcommand{\Qv}{\vc{Q}}
\newcommand{\rv}{\vc{r}}
\newcommand{\sv}{\vc{s}}

\newcommand{\dv}{\vc{d}}
\newcommand{\uv}{\vc{u}}
\newcommand{\Uv}{\vc{U}}

\newcommand{\vv}{\vc{v}}

\newcommand{\phiv}{\vc{\phi}}
\newcommand{\Gammav}{\vc{\Gamma}}

\newcommand{\imag}{\iota}

\newcommand{\Lcal}{\mathcal{L}}
\newcommand{\Ncal}{\mathcal{N}}

\newcommand{\Scal}{\mathcal{S}}

\newcommand{\thetav}{\vc{\theta}}

\newcommand{\Reals}{\mathbb{R}}

\newcommand{\OneVec}{\mathbf{1}}

\usepackage{tikz-qtree}
\usepackage{tikz}
\usepackage{tikz-3dplot}
\usetikzlibrary{calc}
\usetikzlibrary{arrows.meta}
\usetikzlibrary{decorations.pathmorphing}

\usepackage[title]{appendix}

\newcommand{\myreferences}{/Users/fihamze/LaTeXStuff/FirasReferences}

\begin{document}

\title{Parallelized Computation and Backpropagation Under Angle-Parametrized Orthogonal Matrices}

\author{  \textbf{Firas Hamze} \footnote{These authors contributed equally.}\\
  Microsoft Quantum \\
  \texttt{fihamze@microsoft.com}\\
}
\date{\today}

\maketitle 

\begin{abstract}
  We present a methodology for parallel acceleration of learning in
  the presence of matrix orthogonality and unitarity constraints of
  interest in several branches of machine learning.  We show how an
  apparently sequential elementary rotation parametrization can be
  restructured into blocks of commutative operations using a
  well-known tool for coloring the edges of complete graphs, in turn
  widely applied to schedule round-robin (all-against-all) sports
  tournaments. The resulting decomposition admits an algorithm to
  compute a fully-parametrized orthogonal matrix from its rotation
  parameters in $O(n)$ sequential steps and one to compute the
  gradient of a training loss with respect to its parameters in
  $O(n\log n)$ steps. We discuss parametric restrictions of interest
  to generative modeling and present promising performance results
  with a prototype GPU implementation.
\end{abstract}

\section{Introduction}
\label{sec:intro}
We consider in this paper the task of accelerating learning with
generic orthogonal and unitary matrices via parallel
implementation. More specifically, we first observe how a common
parametrization of such matrices of size $n$ as a composition of
$O(n^2)$ elementary operations known as \emph{Givens rotations} can
be optimally restructured into $n-1$ blocks of $n/2$ commutative
transformations which can thus be performed simultaneously. This
representation is then seen to enable a fully-parametrized
orthogonal matrix of size $n$ to be constructively recovered from
its parameters in $O(n)$ sequential steps and the training loss to
be differentiated with respect to its parameters in $O(n\log n)$
steps.

Learning under orthogonality constraints has been an area of
considerable study in recent years: in the training of Recurrent
Neural Networks~(RNNs), it has been noted~\cite{arjovsky2016unitary}
that such weight matrices, being isometric operators, circumvent the
well-known exploding and vanishing gradient problems; in other neural
network contexts, it has been
observed~\cite{harandi2016generalized,ozay2016optimization,huang2018orthogonal}
that imposing orthogonality on weight matrices imparts a beneficial
regularization effect leading to improved generalization. Our reason
for considering the task, not directly connected to either of these
lines, is that such matrices enable precise control over linear
operators in probabilistic generative
models~\cite{rezende2015variational,papamakarios2019normalizing}, to
be discussed in upcoming work.

In all of these settings, learning and optimization under
orthogonality constraints can entail formidable computational costs,
which several approaches have been proposed to reduce. The strategies
depend on the way orthogonal matrices are chosen to be represented;
in~\cite{arjovsky2016unitary}, an efficient parametrization of unitary
matrices in terms of rapidly-computable linear operators is presented,
but was later pointed out in~\cite{wisdom2016full} to be of restricted
capacity, that is only able to represent part of the set of unitary
matrices. These latter authors propose using a parametrization of the
full set of unitary matrices, but the learning procedure involves an
expensive $O(n^3)$ matrix inversion arising due to the Cayley
transform projecting an updated weight matrix onto the Stiefel
manifold, a computational bottleneck. This type of approach was
generalized in~\cite{lezcano2019cheap} where an orthogonal matrix is
approximately (to machine epsilon) represented as the exponential of a
skew-symmetric matrix and learning via gradient descent on the
skew-symmetric set. The algorithm is also $O(n^3)$, which the authors
reasonably justify as being negligible compared to the other
computations involved in training a long RNN; in other settings
however, this cost may be unacceptable when $n$ is large. Furthermore
while in principle feasible, it is unclear how practical it is to
restrict the family of generating skew-symmetric matrices to yield
orthogonal matrices such that rotations in the subspaces spanned by
certain dimensions are excluded, as required in our generative use
case.

In the physics literature, a method proposed by Clements et
al~\cite{clements2016optimal}, improving upon the scheme of Reck et
al~\cite{reck1994experimental}, represents desired unitary matrices via Givens
rotations to yield minimum depth optical interferometers; this
method was used by Jing et al~\cite{jing2017tunable} to propose a tunable
class of unitary RNNs. The motivation behind these physics-based
constructions is quite different from ours; in particular they do
not seek to devise a parallelized representation in which Givens
operators commute; their appeal, as well as those of FFT-inspired
approximation schemes~\cite{mathieu2014fast}, to designing
parametrization of unitaries lies in the fact that they allow rapid
interaction among the coordinates for a given depth and therefore
provide good reduced-capacity approximation classes of arbitrary
unitaries.

In this paper, we do not make approximations and strive to speed up
the computation of and learning with general orthogonal and unitary
matrices. While the main purpose of this work is to expose generic
scope for parallelism when learning with such matrices in principle
accessible to any sufficiently capable hardware, the construction is
well-suited to GPU implementation as it consists almost entirely of
simple operations on independent memory regions, with the exception
of a reduction operation in the gradient evaluation giving rise to
the logarithmic factor in the cost for that step. In addition, like
the previously-cited works involving Givens parametrizations, the
construction allows for reduced-capacity classes, which indeed was
essential in our generative context not as a method for enhancing
generalization but as a means of suitably parametrizing low-rank
dimensionality-reducing (compression) matrices.

In Section~\ref{sec:Preliminaries}, we frame the problem and discuss
the associated computational difficulties; Section~\ref{sec:ForwardU}
discusses how using a crucial construction enabling parallelism, what
we call the round-robin sequence, a straightforward parallelized
algorithm exists for constructing $\Uv$. In Section~\ref{sec:JVP}, the
idea is leveraged to speed up the Jacobian-vector products required
for obtaining the gradients.
Section~\ref{sec:RestrictingParametrization} discusses a type of
parameter-restricted family amenable to the presented
speedups. Section~\ref{sec:Results} presents timing results of a GPU
implementation.  In light of the diverse areas in which this task
arises, this paper will take a somewhat unusual route and focus
exclusively on computational considerations rather than present
accuracy or generalization results under a specific model. Further,
the main paper will focus on real-valued orthogonal matrices;
appropriate steps to extend the ideas to the unitary matrices appear
in the Appendix.

\section{Preliminaries}
\label{sec:Preliminaries}

Consider first the set of \emph{special orthogonal}~(or rotation) matrices
$\{ \Uv \in \SO(n) \}$, that is real matrices such that
$\Uv \Uv^T = \Uv^T\Uv = \Iv$, and $\det \Uv = 1$. It is well-known
that such matrices can be represented by the composition of
$\Ncal \triangleq n(n-1)/2$ Givens rotations in the planes spanned by
all pairs of coordinate axes. In all that follows we assume that $n$
is even for clarity; relaxing this assumption is not difficult. The
order in which these elementary rotations are defined to apply is in
principle arbitrary but needs to be fixed to define a parametrization
of $\SO(n)$; as we will soon see, the chosen order has significant
practical consequences.

Let $E = (e_1, \ldots, e_\Ncal)$ be a sequence of length $\Ncal$ consisting of
pairs of coordinates $e_k = (i,j)$ with $i<j$ such that each pair
appears exactly once; in this paper the $n$ coordinates are assumed
to be labeled with indices in $\{0,\ldots,n-1\}$. The elements of this
sequence define the order in which the Givens rotations are applied.
If $n=4$ and thus $\Ncal=6$, the following are example sequences among
$6!$ possibilities:
\begin{align}
  E = & \big( (0,1), (0,2), (0,3), (1,2), (1,3), (2,3) \big)\nonumber \\
  \tilde{E} = & \big( (0,1), (2,3), (0,2), (1,3), (0,3), (1,2) \big)
\end{align}
where the first element of $E$ is $e_1 = (0,1)$ and so on. Relative
to a chosen $E$, a matrix in $\SO(n)$ can be represented as the product
\begin{align}
  \Uv = \prod_{e \in E} \Gv^e(\theta_e)
  \label{eq:GivensProdDef}
\end{align}
where $\{ \Gv^e(\theta_e)\}$ are the elementary Givens matrices
representing rotation in the span of coordinates $i$ and $j$ defining
$e$ by an angle of $\theta_e$\footnote{Note that in this
  definition $\Gv^{e_\Ncal}$ is the first matrix to apply and
  $\Gv^{e_1}$ is the last.}. The Givens matrices are very sparse
and differ from identity in only $4$ elements; for pair $e = (i,j)$
the differing positions are $(i,i), (j,j), (i,j)$ and $(j,i)$. The
entries of $\Gv^e$ are given by
\begin{align}
  G^e_{ii} & = G^e_{jj} = \cos \theta_{ij} \nonumber \\
  G^e_{ij}  & = -\sin \theta_{ij} \nonumber \\
  G^e_{ji}  & =  \sin \theta_{ij} \nonumber \\
  G^e_{ll}  & = 1  \text{\quad if } l \neq i, l \neq j 
\end{align}
and zero for all other locations. Thus, multiplication by a Givens
matrix only alters rows $i$ and $j$ of the multiplicand.

The angles
$\thetav = (\theta_{e_1}, \ldots \theta_{e_\Ncal}) \in \Reals^\Ncal$
associated with $E$ are the parameters\footnote{Note that
  $\{\theta_e\}$ constructing a specific matrix $\Uv$ relative to two
  composition sequences $E, \tilde{E}$ are in general not the same.}
tracing out the set of matrices in $\SO(n)$.

Real-valued orthogonal matrices of dimension $n$, forming what is
known as the \emph{orthogonal group}, belong to one of two connected
components; the first is the class $\SO(n)$ mentioned above, and the
second is the set of those with determinant -1, sometimes known as
\emph{reflections}. This latter class can easily be parametrized using
the Givens representation by for example negating an arbitrary fixed
column following the construction. The unitary group $\U(n)$ of
complex matrices with $\Uv^\dagger\Uv = \Uv^\dagger\Uv = \Iv_n$ on the
other hand forms a connected space and can be parametrized by the
Givens matrices when the $i^{\textrm{th}}$ column of each $\Gv^e$ is
multiplied by a complex phase factor $e^{\imag\phi_{ij}}$. The choice
of which of these parametrizations is appropriate to a specific
learning task is not within the scope of this work, however we
emphasize that the algorithmic implications of what follows applies to
all of them. To keep the notation to a minimum, we will without loss
of generality focus on $\SO(n)$; relevant adaptations to unitary
matrices are discussed in Appendix~\ref{sec:Appendix:Unitaries}.

In the context of training machine learning
models, two operations are relevant upon inclusion of such a matrix:
\begin{itemize}
\item The \textbf{forward} computation, in which
  $\Uv$ is constructed from its parameters $\thetav$
\item The \textbf{backward} computation, in which the
  \emph{Jacobian-vector product}~(JVP)
  \[
  \frac{\partial\Lcal}{\partial\thetav} = \left(
    \frac{\partial\Uv}{\partial\thetav} \right)^T \frac{\partial\Lcal}{\partial\Uv}
  \]
  is evaluated to determine the gradient of the training loss $\Lcal$
  with respect to the matrix's parameters.
\end{itemize}
These two tasks are not conceptually problematic, however their
\emph{practicality} on large-$n$ systems is not obvious. Consider for
example the forward computation step, which simply corresponds to the
sequence of matrix multiplications defined in
(\ref{eq:GivensProdDef}). The procedure for doing so for a generic
coordinate pair sequence $E$ via a sequence of in-place row operations
performed on an initial identity matrix is described in Algorithm
\ref{alg:GivensFwdGeneral}.
\begin{algorithm}[H]
  \caption{Forward $\Uv$}
  \label{alg:GivensFwdGeneral}
  \begin{algorithmic}[0] 
    \State \textbf{Input:}
    \State \hspace*{\algorithmicindent} Sequence of coordinate pairs $E = (e_1, \ldots, e_\Ncal)$
    \State \hspace*{\algorithmicindent} Parametrizing angles $\thetav = (\theta_{e_1}, \ldots, \theta_{e_\Ncal})$
    \State \textbf{Output:}
    \State \hspace*{\algorithmicindent} $\Uv(\thetav)$
    \State $\Uv \gets \Iv_n$
    \For{$e \in \textrm{reversed}(E)$}
      \State $(i,j) \gets e$
      \State $\rv_i \gets \cos\theta_{ij}\Uv_{i:} -
      \sin\theta_{ij}\Uv_{j:}$
      \State $\rv_j \gets \sin\theta_{ij}\Uv_{i:} +
      \cos\theta_{ij}\Uv_{j:}$
      \State $\Uv_{i:} \gets \rv_i$
      \State $\Uv_{j:} \gets \rv_j$
    \EndFor
  \end{algorithmic}
\end{algorithm}
where for a generic sequence $\sv = ( s_1, \ldots, s_n)$,
$\textrm{reversed}(\sv) \triangleq ( s_n, \ldots, s_1)$. The key point
is that while the operations within the loop body, each of which
jointly implement the in-place application of a Givens rotation on
another matrix, can be performed relatively quickly, taking $O(n)$
sequential operations and reducible to constant time given sufficient
parallel resources, there are $\Ncal = O(n^2)$ such applications to
perform and in general these products do not commute. In other words,
a certain application order must be respected, and so the construction
of $\Uv$ from its angles appears to be an obligately \emph{sequential}
task of $O(n^2)$ steps, each comprising $O(n)$ operations on
independent pairs of elements. Even assuming parallelized
implementation of the steps within the loop, the overall $O(n^2)$ cost
of determining $\Uv$ from $\thetav$ becomes a serious issue for
realistic-sized $n$.

Closer inspection will reveal however that there is indeed
considerable scope for parallelism; this will be detailed in the
following sections, but the idea is to construct the sequence $E$ in a
careful manner, namely such that it consists of $n-1$ subsequences, or
computational blocks, of $n/2$ elements each, with the property that
the Givens rotations within each block apply to \emph{disjoint pairs}
of coordinates and hence commute within the block. This fortunate
property therefore allows $n/2$ Givens rotations to be applied
simultaneously and reduce the complexity of the forward computation of
$\Uv$ to $O(n)$ sequential steps, each of which can ideally take
constant time. The property can also be exploited to yield a
substantial performance gain on the JVP computation; in particular we
will present a method for backpropagation with respect to the
$\thetav$ in $O(n\log n)$ sequential steps, where the logarithmic
factor represents the parallelized complexity of reduction operations
arising in the method.

\section{Forward $\Uv$ Computation via Round-Robin Sequences}
\label{sec:ForwardU}

Suppose we seek to arrange the coordinate pairs into the smallest
number of subsequences (blocks) such that the following properties
hold:
\begin{itemize}
\item Within a block, no two pairs share a coordinate
\item Each pair appears in exactly one block
\end{itemize}
For reasons that we will discuss shortly we refer to coordinate pair
sequences satisfying these properties as \emph{round-robin
  sequences}. We notice that this task is equivalent to the
graph-theoretic problem of finding an optimal \emph{edge-coloring}, in
this case of the complete graph of $n$ variables $K_n$; more
specifically, given a complete graph the objective is to assign a
minimum number of color labels to all the edges such that no edges
incident on a vertex have the same color. This problem and its
generalizations are well-studied~\cite{baranyai1974factorization}, and
it is well-established that for even $n$ the goal is achieved on $K_n$
using $n-1$ colors. Clearly, our coordinate pairs $e = (i,j)$ play the
role of edges, the coordinates themselves are the vertices, and the
colors index the computational blocks to which the pairs are assigned.

An example of a sequence exhibiting the desired properties for $n=6$ is
\begin{align}
  E = ( \overbrace{ (0,5) (1,4) (2,3) }^{b_1}, \overbrace{ (0,4) (3,5) (1,2)}^{b_2}, \overbrace{ (0,3) (2,4) (1,5) }^{b_3}, \overbrace{ (0,2) (1,3) (4,5) }^{b_4},
  \overbrace{ (0,1) (2,5) (3,4) }^{b_5} )
  \label{eq:exampleRRSeq}
\end{align}
where the $n-1=5$ blocks $\{ b_1, \ldots, b_5 \}$ each of size
$n/2=3$, are indicated. The principal algorithmic advantage of such a
parameter ordering is that within each block, the updates to the
matrix take place over \emph{independent} pairs of rows, are
applicable in arbitrary order, and can hence be fully parallelized. To
make this more explicit, if we define the sequence of \emph{blocks}
with $B = (b_1, b_2, \ldots, b_{n-1} )$ we can re-express the generic
in-place Algorithm \ref{alg:GivensFwdGeneral} into Algorithm
\ref{alg:GivensFwdRR}, its round-robin parallel variant.
\begin{algorithm}[H]
  \caption{Parallel Forward $\Uv$}
  \label{alg:GivensFwdRR}
  \begin{algorithmic}[0] 
    \State \textbf{Input:}
    \State \hspace*{\algorithmicindent} Block sequence $B = ( b_1,
    \ldots, b_{n-1})$ such that $E$ is round-robin
    \State \hspace*{\algorithmicindent} Parametrizing angles $\thetav = (\theta_{e_1}, \ldots, \theta_{e_\Ncal})$
    \State \textbf{Output:}
    \State \hspace*{\algorithmicindent} $\Uv(\thetav)$
    \State $\Uv \gets \Iv_n$
    \For{$b \in \textrm{reversed}(B)$}
      \For{$e \in b$} \Comment{This block is fully parallel}
        \State $(i,j) \gets e$
        \State $\rv_i \gets \cos\theta_{ij}\Uv_{i:} -
        \sin\theta_{ij}\Uv_{j:}$
        \State $\rv_j \gets \sin\theta_{ij}\Uv_{i:} +
        \cos\theta_{ij}\Uv_{j:}$
        \State $\Uv_{i:} \gets \rv_i$
        \State $\Uv_{j:} \gets \rv_j$
      \EndFor
      \State Synchronize parallel operations
    \EndFor
  \end{algorithmic}
\end{algorithm}
The synchronization step appearing following the inner loop is to
emphasize that for correct behavior, the parallel block operations
must fully complete before proceeding to the next block. The
sequential step complexity is clearly $O(n)$. It should be mentioned
that this statement assumes that relevant worker threads have
concurrent read access to the values of $\theta_{ij}$. If this does
not hold, the time to broadcast each $\theta_{ij}$, typically
logarithmic in $n$ for each parallel block, must be taken into
account.

\begin{figure}
\[
\xymatrix{
 *+[Fo]{0}\ar@/{^1.5pc}/@{-}[rrrrr] &
 1 \ar@/{^1pc}/@{-}[rrr] &
 2 \ar@/{^0.5pc}/@{-}[r] &
 3 &
 4 &
 5 &
 b_1 =  (0,5) (1,4) (2,3)
}
\]
\vspace{0.25cm}
\[
\xymatrix{
 *+[Fo]{0}\ar@/{^1.5pc}/@{-}[rrrrr] &
 5 \ar@/{^1pc}/@{-}[rrr] &
 1 \ar@/{^0.5pc}/@{-}[r] &
 2 &
 3 &
 4 &
 b_2 = (0,4) (3,5) (1,2)
}
\]
\vspace{0.25cm}
\[
\xymatrix{
 *+[Fo]{0}\ar@/{^1.5pc}/@{-}[rrrrr] &
 4 \ar@/{^1pc}/@{-}[rrr] &
 5 \ar@/{^0.5pc}/@{-}[r] &
 1 &
 2 &
 3 &
 b_3 = (0,3) (2,4) (1,5)
}
\]
\vspace{0.25cm}
\[
\xymatrix{
 *+[Fo]{0}\ar@/{^1.5pc}/@{-}[rrrrr] &
 3 \ar@/{^1pc}/@{-}[rrr] &
 4 \ar@/{^0.5pc}/@{-}[r] &
 5 &
 1 &
 2 &
 b_4 = (0,2) (1,3) (4,5)
}
\]
\vspace{0.25cm}
\[
\xymatrix{
 *+[Fo]{0}\ar@/{^1.5pc}/@{-}[rrrrr] &
 2 \ar@/{^1pc}/@{-}[rrr] &
 3 \ar@/{^0.5pc}/@{-}[r] &
 4 &
 5 &
 1 &
 b_5 = (0,1) (2,5) (3,4)
}
\]
\caption{An illustration of the \emph{circle method} for generating
  the round-robin sequence appearing in
  Equation~(\ref{eq:exampleRRSeq}). Each step, corresponding to a horizontal
  line, is associated with a sequence of dimensions (the left list of
  numbers).  At a step, the pairing (indicated by arcs) among
  numbers in the left sequence at equal distance from the endpoints
  define the pairs of coordinates within the step's block (at right);
  these coordinates can be Givens-rotated independently and in
  parallel. The next step's dimension sequence is obtained by shifting
  modulo $n-1$ the last $n-1$ elements of the current step's while
  holding the first element (0 in this case) fixed. The initial
  dimension sequence was here taken to be $(0,1,\ldots,5)$, but could
  have been arbitrarily permuted to yield a different round-robin
  sequence. The result is an enumeration of all coordinate pairs $\{(i,j):i<j\}$
  in $n-1=5$ sequential steps such that each step no pairs share a
  coordinate.}
\label{fig:circleAlgEx}
\end{figure}
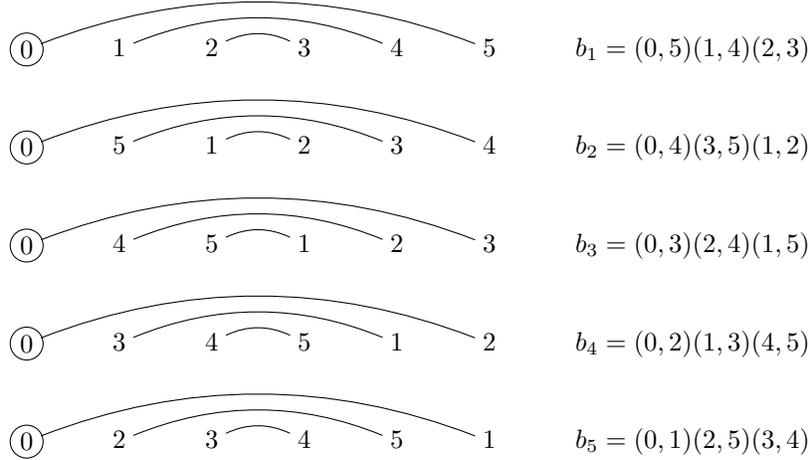

Of course we have not yet discussed how to construct round-robin
sequences for general (even) $n$. Fortunately, there are well-known
methods for such construction in the context of another equivalent
problem: that of scheduling \emph{round-robin} sports tournaments, in
which all teams must compete against each other in as few rounds of
concurrently occurring games as possible. The specific procedure we
use has been known since the mid-19th
century~\cite{kirkman1847problem}, is commonly called the \emph{circle
  method} today, and is widely employed for tournament scheduling.
Rather than formally describe it, the steps used to generate the
example sequence (\ref{eq:exampleRRSeq}) are shown in
Figure~\ref{fig:circleAlgEx}. The main idea is as follows.  Beginning
with an arbitrary permutation of the coordinate sequence
$(0,\ldots,n-1)$, in this case $(0, \ldots, 5)$, the first block is
obtained by pairing coordinates at the same distance from the
endpoints (illustrated with the arcs). Then, holding one element
fixed, in this case the $0$ coordinate, we perform $n-2$ shifts modulo
$n-1$ on the remaining elements; at each shift, a block is again
obtained by pairing coordinates equally-spaced from the ends.

While we have assumed that $n$ was even, extending the algorithm to
odd $n$ is straightforward at modest cost: in constructing the round
robin sequence, we first augment the initial dimension sequence to
being some permutation of $(0, \ldots, n)$ instead of $(0,\ldots,n-1)$
and construct the blocks as already described. Since the added
dimension is not actually part of the model, we complete the
adaptation to odd $n$ by adding a condition to bypass the parallel
loop in Algorithm \ref{alg:GivensFwdRR} if $j = n$.

In our implementation of Algorithm \ref{alg:GivensFwdRR}, the inner
loop was performed by a CUDA~\cite{cuda} kernel, which from the
mathematical structure of the round-robin construction, is able to
determine the relevant $i$ and $j$ under consideration based on the
passed-in current step in the outer loop. In other words, the
round-robin sequence does \emph{not} need to be explicitly sent to the
processing device but can be independently and lazily determined by
the computational threads when they are informed of the current
sequential step. The overall host/device data transfer requirements of
the algorithm are minimal.

\section{Backpropagation through $\Uv$}
\label{sec:JVP}

We now turn our attention to the task of efficiently computing the JVP
associated with the gradient of a training loss with respect to the
orthogonal matrix's parameters $\thetav$. The path to this goal is
somewhat more involved than that to the forward construction, but the
result is a remarkably simple algorithm; unsurprisingly, its
asymptotic (in number of processors) complexity is higher than that of
the forward operation but at $O(n\log n)$ is still relatively modest
considering the task at hand. In the following section, we proceed to
analytically characterize the Jacobian of $\Uv$; the resultant
properties will be used in Section~\ref{sec:JVP:ComputingGrad} for the
parallel calculation of the JVP in which we are ultimately interested.
As is often the case when training deep models, the Jacobian itself
need never be computed explicitly.

\subsection{The Jacobian of $\Uv$ Under Round-Robin Parallelism}
\label{sec:JVP:Jacobian}

We presently consider the determination of
\[
\frac{\partial\Uv}{\partial\theta_e}
\]
for round-robin structured coordinate pairs $e \in E$; when reshaped
into a vector, these objects form the columns of the $n^2 \times \Ncal$
Jacobian matrix.

Note first that
\[
\Gv^e{}' \triangleq \frac{\partial\Gv^e}{\partial\theta_e}
\]
is a matrix consisting of only $4$ nonzero elements
\begin{align}
  \big[ \Gv^e{}' \big]_{ii} = & \big[ \Gv^e{}' \big]_{jj} = -\sin\theta_e \nonumber \\
  \big[ \Gv^e{}' \big]_{ij} = & -\cos\theta_e \nonumber \\
  \big[ \Gv^e{}' \big]_{ji} = & \cos\theta_e
\end{align}
The product
\begin{align}
  \Qv_e \triangleq \Gv^e{}' \Gv^e{}^T
  \label{eq:Qdef}
\end{align}
can then be seen to be nonzero only at positions $(i,j)$ and $(j,i)$
where it is $-1$ and $1$ respectively.

Let us express $\Uv$ as a product of $n-1$ compositions of Givens
rotations over the blocks in $B$:
\[
\Uv = \prod_{b \in B} \Gv^b(\thetav_b)
\]
where
\[
\Gv^b(\thetav_b) = \prod_{e\in b}\Gv^e(\theta_e)
\]
and $\thetav_b$ is the subset of the angles defined by the block.

If $e \in b_k$ then
\begin{align}
\frac{\partial\Uv}{\partial\theta_e} = \Gv^{b_1}\ldots
  \frac{\partial \Gv^{b_k}}{\partial\theta_e}\ldots \Gv^{b_{n-1}}
  \label{eq:UJacBlocks}
\end{align}
But by the commutativity \emph{within} blocks, for any $\tilde{e} \in b_k$
\begin{align}
  \Gv^{b_k}(\thetav_{b_k}) = & \prod_{e\in {b_k}}\Gv^e(\theta_e) \nonumber \\
  = & \Gv^{\tilde{e}}(\theta_{\tilde{e}})\prod_{\substack{e\in b_k\\e\neq \tilde{e}}}\Gv^e(\theta_e)
\end{align}

Hence
\begin{align}
  \frac{\partial \Gv^{b_k}}{\partial\theta_{\tilde{e}}} = & \frac{\partial\Gv^{\tilde{e}}}{\partial\theta_{\tilde{e}}} \prod_{\substack{e\in b_k\\e\neq \tilde{e}}}\Gv^e \nonumber \\
                                                         = & \frac{\partial\Gv^{\tilde{e}}}{\partial\theta_{\tilde{e}}} \Gv^{\tilde{e}}{}^T \Gv^{b_k}
\end{align}
where the second equality expresses removal of the rotation
corresponding to $\tilde{e}$ from the block's joint operation.

Using the definition (\ref{eq:Qdef}) we therefore have for $\tilde{e}
\in b_k$
\begin{align}
\frac{\partial \Gv^{b_k}}{\partial\theta_{\tilde{e}}} =
  \Qv_{\tilde{e}}\Gv^{b_k}
  \label{eq:blockQPartial}
\end{align}

This will be used to recursively ``evaluate''
$\frac{\partial \Uv}{\partial\theta_e}$ for all $\theta_e$; the
quotation marks are to again emphasize that we do not literally carry
out this computation, but need its form to efficiently define the JVP.
We define the running matrix products:
\begin{align}
  \Uv^{1:k} \triangleq & \prod_{b \in b_1,\ldots,b_k}\Gv^b \nonumber \\
  \Uv^{k:n-1} \triangleq & \prod_{b \in b_k,\ldots,b_{n-1}}\Gv^b
\end{align}
From~(\ref{eq:UJacBlocks}) and~(\ref{eq:blockQPartial}) we thus have
for $e \in b_k$
\[
\frac{\partial\Uv}{\partial\theta_e} = \Uv^{1:k-1} \Qv_e \Uv^{k:n-1} 
\]

An essential aspect in the parallel computation of
$\frac{\partial\Uv}{\partial\theta_e}$ for all $e$ in a block is the
fact that in-place updates of $\Uv^{1:k-1}$ and $\Uv^{k:n-1}$ can be
readily parallelized from one step to the next. More precisely, let
$\Uv^\fwd$ and $\Uv^\bck$ be matrices storing, respectively,
$\Uv^{1:k-1}$ and $\Uv^{k:n-1}$ at sequential step $k$; these matrices
can be quickly modified from their values at step $k+1$, when they
represented $\Uv^{1:k}$ and $\Uv^{k+1:n-1}$ respectively. This is
because updating $\Uv^\bck$ involves \emph{pre-multiplying} it by
$\Gv^{b_k}$ to include the effect of the rotations in the block while
updating $\Uv^\fwd$ involves \emph{post-multiplying} it by
$\Gv^{b_k}{}^T$ to remove the effect of the block's rotations. Due to
the round-robin imposed independence of the coordinate pairs within a
block, these operations can be performed in parallel on the relevant
rows of $\Uv^\bck$ and columns of $\Uv^\fwd$. When $\Uv^\fwd$ is
initialized to $\Uv$ as determined in the forward computation and
$\Uv^\bck$ is set to identity, the parallel procedure for their
updates takes place in completely analogous fashion to that of the
forward computation.

We are finally ready to describe evaluating the derivative of $\Uv$; 
having suitably updated $\Uv^\fwd$ and $\Uv^\bck$ at step $k$, 
we require for all $e \in b_k$
\begin{align}
  \frac{\partial\Uv}{\partial\theta_e} = \Uv^\fwd \Qv_e \Uv^\bck
  \label{eq:delUvFwdBck}
\end{align}
Let
\[
\Uv^\fwd = \begin{bmatrix}
    \uv_1 & \uv_2 & \hdots & \uv_n
\end{bmatrix}
\]
and
\[
\Uv^\bck = \begin{bmatrix}
    \vv^T_1 \\
    \vv^T_2 \\
    \vdots \\
    \vv^T_n
\end{bmatrix}
\]
From the fact that $\Qv_e$ is $-1$ at $(i,j)$, $1$ at $(j,i)$, and
zero everywhere else it can be shown that (\ref{eq:delUvFwdBck})
reduces to
\begin{align}
  \frac{\partial\Uv}{\partial\theta_e} = \uv_j\vv_i^T - \uv_i\vv_j^T
  \label{eq:delUvdelThetae}
\end{align}
In other words, the derivatives of $\Uv$ with respect to the
coordinate pairs within a block are rank-2 matrices constructed from
independent columns and rows of $\Uv^\fwd$ and $\Uv^\bck$
respectively, which are \emph{fixed for the block}. Hence if we were
actually interested in computing the Jacobian of $\Uv$ the ``columns''
corresponding to $\thetav_{b_k}$would in principle be computable in
parallel. It turns out that the structure we have exposed carries over
to the more relevant problem of JVP computation, to which we now turn.

\subsection{Computing the Gradient}
\label{sec:JVP:ComputingGrad}

The automatic differentiation procedure is assumed to provide the
gradient of the loss with respect to the outputs of the function
$\Uv(\thetav)$, here assumed to be structured as an $n \times n$
matrix
\[
\Gammav \triangleq \frac{\partial\Lcal}{\partial\Uv}  =
\begin{bmatrix}
  \gv_1^T \\
  \gv_2^T \\
  \vdots \\
  \gv_n^T
  \end{bmatrix}
\]

Suppose now that we are at block $b_k$; let
\[
\frac{\partial\Uv}{\partial\theta_e} = \begin{bmatrix}
  \cv^e_1{}^T \\
  \cv^e_2{}^T \\
  \vdots \\
  \cv^e_n{}^T
  \end{bmatrix}
\]
From the property (\ref{eq:delUvdelThetae}) obtained in the previous
section, we see that the $l^{\textrm{th}}$ row vector of the partial matrix is given
by
\[
\cv_l^e{}^T = u_{jl} \vv_i^T - u_{il}\vv_j^T
\]

For the purpose of the deriving the JVP it is convenient to conceptually
reshape the matrix $\Uv$ into a vector whose elements correspond
to those of $\Uv$ in row-major order. In that representation, the
\emph{column} of the Jacobian
\[
\frac{\partial\Uv}{\partial\thetav}
\]
corresponding to $\theta_e$ is simply the length $n^2$ vector
\[
\begin{bmatrix}
  \cv^e_1 \\
  \cv^e_2 \\
  \vdots \\
  \cv^e_n
\end{bmatrix}
\]
while $\Gammav$ is reshaped to
\[
\begin{bmatrix}
  \gv_1 \\
  \gv_2 \\
  \vdots \\
  \gv_n
\end{bmatrix}
\]
so that the JVP yields the following loss gradient component with
respect to $\theta_e$:
\begin{align}
\frac{\partial\Lcal}{\partial\theta_e} = & \begin{bmatrix}
  \cv^e_1{}^T &
  \cv^e_2{}^T &
  \hdots &
  \cv^e_n{}^T
\end{bmatrix}
\begin{bmatrix}
  \gv_1 \\
  \gv_2 \\
  \vdots \\
  \gv_n
\end{bmatrix} \nonumber \\
  = & \sum_{l=1}^nu_{jl}\vv_i^T\gv_l  - \sum_{l=1}^nu_{il}\vv_j^T\gv_l
      \label{eq:delLdelThetae1}
\end{align}
Finally, defining the matrix $\Mv$ to be
\begin{align}
  \Mv \triangleq & \begin{bmatrix}
    \vv_1^T \\
    \vv_2^T \\
    \vdots \\
    \vv_n^T
  \end{bmatrix}
  \begin{bmatrix}
    \gv_1 & \gv_2 & \hdots & \gv_n 
  \end{bmatrix} \nonumber \\
  & = \Uv^\bck \Gammav^T
    \label{eq:MDef}
\end{align}
we rewrite (\ref{eq:delLdelThetae1}) to obtain the key expression for
the gradient of $\Lcal$ with respect to $\theta_e$ with $e = (i,j)$
\begin{align}
  \frac{\partial\Lcal}{\partial\theta_e} = \Mv_{i:}\uv_j -
  \Mv_{j:}\uv_i
  \label{eq:delLdelThetaeFinal}
\end{align}

We now discuss parallel evaluation of these gradient components for
all $e \in b_k$. In the previous section, we discussed how $\Uv^\fwd$
and $\Uv^\bck$ could both be efficiently modified in place over the
sequence of blocks. In the context of computing the JVP however, we
observe that while $\Uv^\fwd$ is still required (its columns
$\{ \uv_l \}$ appear in~(\ref{eq:delLdelThetaeFinal})), we now longer
require explicit access to $\Uv^\bck$, but only to the matrix $\Mv$
corresponding the product of $\Uv^\bck$ with the upstream loss
gradient $\Gammav$ (transposed). Fortunately it is also efficient to
compute using precisely the same idea employed for $\Uv^\bck$ itself;
from the definition of $\Mv$ it is apparent that the only required
modification is to commence the update recursion with $\Gammav^T$
instead of an identity matrix.

Once $\Mv$ and $\Uv^\fwd$ have been updated for block $b_k$, the
parallel method for simultaneously evaluating
$\frac{\partial\Lcal}{\partial\theta_e}$ for $e \in b_k$ can be fully
described. Suppose $\Av$ is a temporary $n/2 \times n $ matrix, which
need only be allocated once at the outset. Each coordinate pair
$e \in b_k$ is mapped to a row of $\Av$. Let $m(e)$ define this
mapping, an example of which in a GPU context is the index in the
computational grid of the block processing coordinate pair $e$.
Parallel threads first assign the rows of $\Av$ such that if
coordinate pair $e$ maps to row $m$, then for $l \in \{1,\ldots,n\}$
\[
A_{ml} \gets M_{il}u_{lj} - M_{jl}u_{li}
\]
Note that this computation and assignment again costs a small,
constant number of arithmetic operations and requires accesses to
independent memory regions. The gradient of the loss with respect to
$\thetav_{b_k}$ is finally obtained via the \emph{reduction} operation
of summing the rows of $\Av$, or equivalently multiplying $\Av$ by a
vector of ones, the result of which is assigned to the relevant
storage of the gradient.
\begin{align}
  \dv \gets & \Av\OneVec_n \nonumber \\
  \frac{\partial\Lcal}{\partial\theta_e} \gets & d_{m(e)}
\end{align}
This reduction step is the only component of either the forward or
backward algorithms in which the strict independence of memory regions
accessed by parallel threads is broken. Consisting of $n/2$ parallel
summations of length $n$ vectors, its parallel step complexity is
$O(\log n)$, and eliciting its best practical performance takes
considerable care. Fortunately, this ubiquitous task has been the
subject of much thought and
craftsmanship(e.g.~\cite{harris2007optimizing}) and has inspired
optimized implementations(e.g.~\cite{cublas}) on which we can
rely.

The overall procedure is presented in Algorithm~\ref{alg:JVPRR}, from
which it can be seen that the sequential iteration over the $n-1$
blocks yields a parallel time of $O(n\log n)$. We note that extension
to odd $n$ is straightforward by augmenting with an inactive dimension,
analogously to how it was described in the forward stage.
\begin{algorithm}[H]
  \caption{Parallel JVP}
  \label{alg:JVPRR}
  \begin{algorithmic}[0] 
    \State \textbf{Input:}
    \State \hspace*{\algorithmicindent} Block sequence $B = ( b_1, \ldots, b_{n-1})$ such that $E$ is round-robin
    \State \hspace*{\algorithmicindent} Parametrizing angles $\thetav = (\theta_{e_1}, \ldots, \theta_{e_\Ncal})$
    \State \hspace*{\algorithmicindent} $\Uv$ computed by Forward Algorithm~\ref{alg:GivensFwdRR}
    \State \hspace*{\algorithmicindent} Loss gradient with respect to outputs $\Gammav$
    \State \textbf{Output:}
    \State \hspace*{\algorithmicindent} Loss gradient with respect to parameters: $ \frac{\partial\Lcal}{\partial\thetav} $
    \State \textbf{Initialize:}
    \State \hspace*{\algorithmicindent} $\Uv^\fwd \gets \Uv$
    \State \hspace*{\algorithmicindent} $\Mv \gets \Gammav^T $
    \State \hspace*{\algorithmicindent} $\Av \gets$ empty $N/2 \times N$ matrix
    \For{$b \in \textrm{reversed}(B)$} 
      \For{$e \in b$} \Comment{Parallel $\Uv^\fwd$ update: $O(1)$}
        \State $(i,j) \gets e$
        \State $\cv_i \gets \cos\theta_{ij}\Uv^\fwd_{:i} - \sin\theta_{ij}\Uv^\fwd_{:j}$
        \State $\cv_j \gets \sin\theta_{ij}\Uv^\fwd_{:i} + \cos\theta_{ij}\Uv^\fwd_{:j}$
        \State $\Uv^\fwd_{:i} \gets \cv_i$
        \State $\Uv^\fwd_{:j} \gets \cv_j$
      \EndFor
      \State Synchronize
      \For{$e \in b$} \Comment{Parallel $\Mv$ update: $O(1)$}
        \State $(i,j) \gets e$
        \State $\rv_i \gets \cos\theta_{ij}\Mv_{i:} - \sin\theta_{ij}\Mv_{j:}$
        \State $\rv_j \gets \sin\theta_{ij}\Mv_{i:} + \cos\theta_{ij}\Mv_{j:}$
        \State $\Mv_{i,:} \gets \rv_i$ 
        \State $\Mv_{j,:} \gets \rv_j$
      \EndFor
      \State Synchronize
      \For{$e \in b$}  \Comment{Parallel $\Av$ assignment: $O(1)$}
        \State $(i,j) \gets e$
        \State $m \gets m(e)$
        \For{$l \in \{0, \ldots, n-1\}$}  \Comment{Parallel}
          \State $A_{ml} \gets M_{il}u_{lj} - M_{jl}u_{li}$
        \EndFor
      \EndFor
      \State Synchronize
      \State $\dv \gets \Av \OneVec_n$ \Comment{Sum rows of $\Av$: $O(\log n)$}
      \For{$e \in b$}  \Comment{Parallel JVP assignment: $O(1)$}
        \State $(i,j) \gets e$
        \State $m \gets m(e)$
        \State $\frac{\partial\Lcal}{\partial\theta_e} \gets d_{m}$
      \EndFor
    \EndFor
  \end{algorithmic}
\end{algorithm}

\section{Restricting the Parametrization}
\label{sec:RestrictingParametrization}

For some tasks, e.g. dimensionality reduction and generative modeling,
it is necessary to learn over certain restrictions of the class of
orthogonal matrices. Of particular interest is the ability to specify
a subset of coordinates such that no rotations corresponding to the
subset's coordinate pairs are performed. Without loss of generality,
for integer $m \leq n-1$ let $\Scal = (0, \ldots, m-1 )$,
$\bar{\Scal} = (m,\ldots,n-1)$, and $\kappa \triangleq n-m$. We seek
to represent the orthogonal matrices representable by all pairs of
Givens rotations \emph{excluding} those corresponding to the
$\kappa(\kappa-1)/2$ pairs in $\bar{\Scal}$, or equivalently, to only
apply the $\Ncal = mn - m(m+1)/2$ rotations equivalent to the
remaining pairs. For example if $n=8$ and $m=4$, the pairs
$\{(4,5), (4,6), (4,7), (5,6), (5,7), (6,7) \}$ are removed from the
full set of $28$ pairs leaving $22$ free parameters. This restriction
can be used as a tool to parametrize arbitrary $m \times n$ matrices.

Determining an optimal blocking strategy striving to partition the
pairs of interest into a minimum number of commutative blocks is still
equivalent to finding an optimal edge-coloring, but now rather than
being over the complete graph of $n$ nodes, it is over a graph in
which the first $m$ nodes are adjacent to all other nodes and the last
$\kappa$ nodes are not connected to each other. We are not aware of
prescriptions analogous to the circle method for constructing optimal
colorings for this class of graphs, but we make the rather obvious
point that the algorithm for the unrestricted case can still be used
to yield $n-1$ blocks, now no longer of size $n/2$. To effect this,
the construction and gradient algorithms are adapted by augmenting the
parallel loops with a condition to bypass all operations if
$i \geq m$; by convention $i<j$ so the bypass condition is met when
$(i,j)$ is an excluded pair and that consequently $\theta_{ij}$ is not
a free parameter. Note that this does not increase the arithmetic
operation count over the purely sequential computation. Further, if
$m$ scales linearly with $n$, that is $m = \alpha n $ for some fixed
$\alpha \in (0,1]$, the total number of pairs $\Ncal$ remains $O(n^2)$
and hence the parallel methods under this subspace-restricted block
construction continue to yield the stated asymptotic speedups
discussed earlier.

\section{Results}
\label{sec:Results}
\begin{figure*}
  \centering
  \includegraphics[width=0.45\columnwidth]{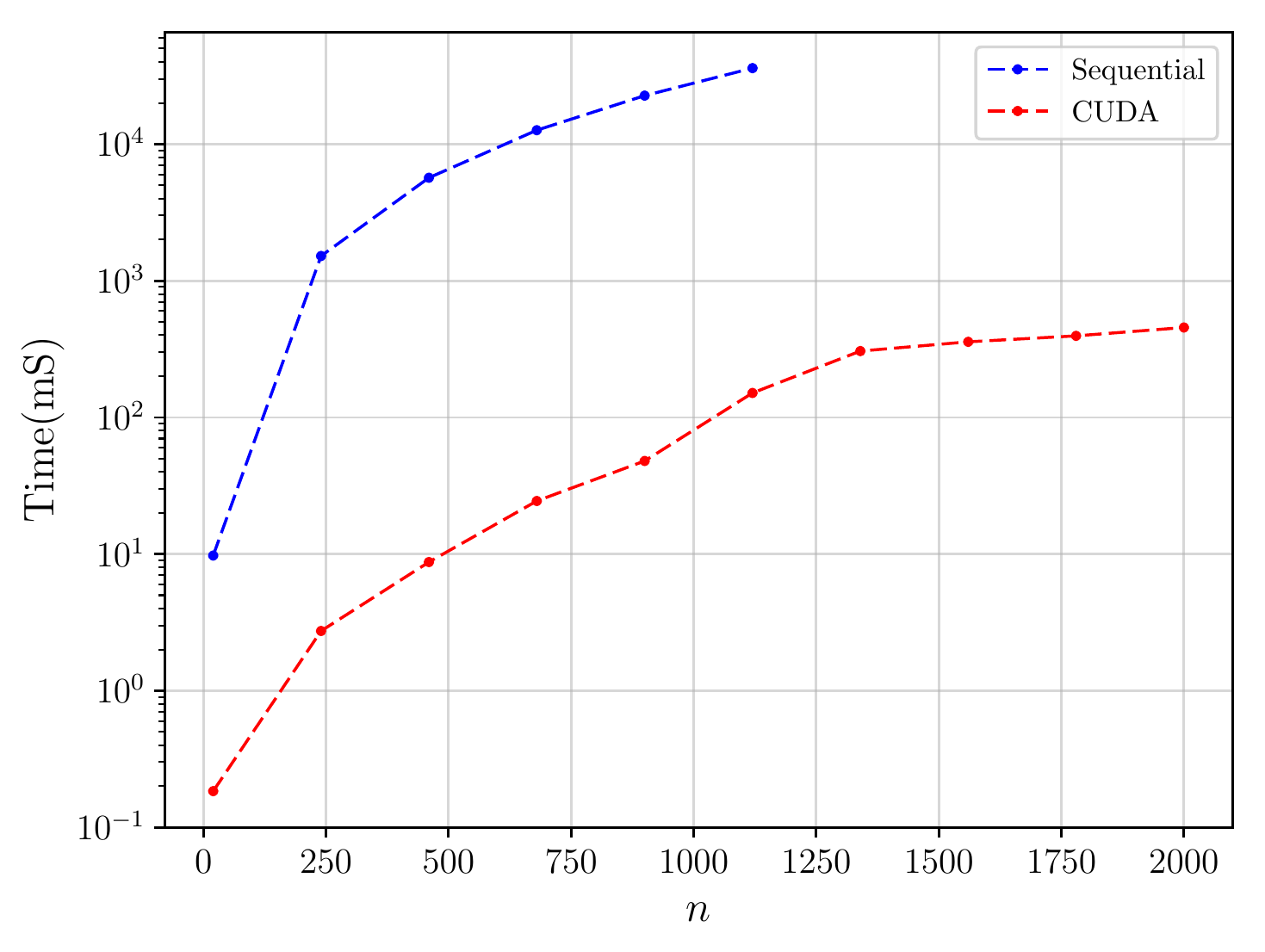}
  \includegraphics[width=0.45\columnwidth]{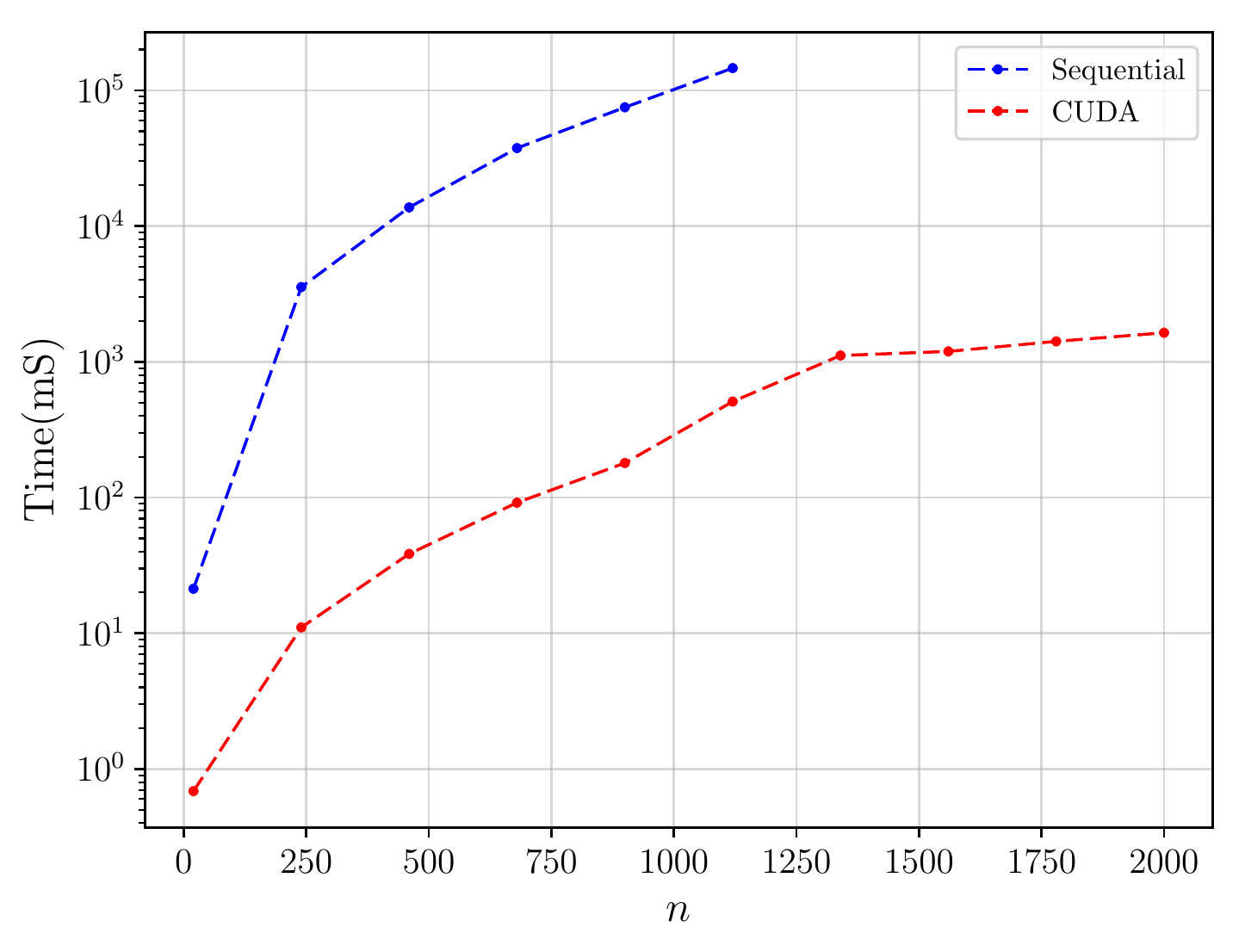}
  \caption{Timing results (in milliseconds) for sequential versus
    GPU-parallel computation of the forward construction (left) and
    gradient computation (right) with respect to orthogonal matrices
    whose size is on the $x$-axes. See text for discussion.}
  \label{fig:resultsTiming}
\end{figure*}

This section will present numerical results illustrating the viability
of the proposed parallelizations. Prior to turning to the details, a
few remarks are in order. First, we emphasize that while we have
opted for a GPU-centered implementation, the main objective of this
work is not to advocate a particular computing paradigm but rather to
generically expose the inherent parallelizability of forward and
backpropagation under angle-parametrized orthogonal matrices. GPUs
are appealing due to their suitability to processing large numbers of
relatively independent operations such as those present in our
round-robin block matrix updates, but in principle our construction
can be leveraged in any parallel computing context, e.g. multi-core
CPUs, etc.

Second, it is not within the scope of this paper to exhaustively run
the gamut of low-level optimization techniques to tease out the best
possible performance; though clearly essential for practical adoption
of the algorithm at scale, this will be left to later work. Our
current proof-of-concept aim is more modest: we will compare the times
taken by a sequential but reasonably efficient single-threaded
CPU-centered implementation against a GPU variant instantiating the
ideas in this paper. The experiments serve mostly as preliminary
evidence that the promised parallel speedups appear to be real and not
hindered by some unforeseen roadblock.

Both algorithms were implemented as extensions to the PyTorch
Library~\cite{paszke2019pytorch}. The CPU version was written in C++,
striving for efficiency by leveraging optimized tensor operations
implemented in PyTorch's ATen library when possible, and needless to
say, by exploiting the sparsity structure of Givens rotations (i.e. a
Givens matrix only acted on the relevant pair of rows of the operand
and was most certainly not implemented as multiplication by a full
$n \times n$ matrix). The GPU variant implemented the parallel
operations with customized CUDA kernels (except for the backward
step's reduction, which used ATen's summation function) and were
called from sequential C++ code iterating over the $n-1$ blocks
partitioning the round-robin sequence.

The CPU code was run on an Intel Xeon E5-2690 processor while the GPU
runs used an Nvidia Tesla K80, hardly a state-of-the-art device at the
time of writing; our aim was to measure the average time (in
milliseconds) taken by each device over 50 runs of the forward and
backward algorithms as a function of matrix dimension $n$. Due to the
excessive time consumed by the backward stage, the sequential method
only considered matrices up to size $n=1120$ while those on the GPU
were allowed to be as large as $n=2000$. The timing for the GPU did
not consider the latency of uploading parameters $\thetav$ onto the
device, which should only occur once in a training context.

Results of the computations are shown in
Figure~\ref{fig:resultsTiming} with the forward and backward times
appearing in the left and right plots respectively, where we see that
for our most likely suboptimized CUDA implementation, the speedups
for both stages using the parallelizations described in this paper
measure in the hundreds, strongly suggesting that a serious
undertaking to engineer the code is in order, and may have positive
impact on the efficiency of training RNNs and broad classes of
generative models used in machine learning today.

\section*{Acknowledgements}
FH is indebted to Stephen Jordan, Rishit Sheth, and Brad Lackey for valuable
discussions and feedback.

\bibliography{\myreferences}

\begin{appendices}
\section{Adaptations to $\U(n)$}
\label{sec:Appendix:Unitaries}

Extension of this paper's parallelization to the group of unitary
matrices $\U(n)$ is in principle straightforward; extra costs arise
due to the need to handle the complex phase parameters but the overall
complexity is essentially unchanged. In this appendix, we describe the
extended parametrization involving complex phase
factors(e.g.~\cite{clements2016optimal}), present the constructive
algorithm, and describe computation of the derivatives within a block
with respect to both the rotation and the phase angle parameters.

Given a coordinate pair sequence $E$, an elementary rotation
associated with pair $e = (i,j)$ with $i<j$ is $\Gv^e(\theta_e, \phi_e)$, where
\begin{align}
  G^e_{ii} & = G^e_{jj} = e^{\imag\phi_{ij}}\cos \theta_{ij} \nonumber \\
  G^e_{ij}  & = -\sin \theta_{ij} \nonumber \\
  G^e_{ji}  & =  e^{\imag\phi_{ij}}\sin \theta_{ij} \nonumber \\
  G^e_{ll}  & = 1  \text{\quad if } l \neq i, l \neq j 
\end{align}
and zero everywhere else; the \emph{phase} parameters
$\phiv = (\phi_{e_1}, \ldots, \phi_{e_\Ncal})$ now contribute complex
\emph{phase factors} $\{ e^{\imag\phi_e} \}$ to the rotations. It
should be clear that for given round-robin $E$ and parameters
$\thetav, \phiv$, the structure of the parallel forward construction
of $\Uv$ as defined in Equation~\ref{eq:GivensProdDef} is unaltered;
for completeness, it is presented in
Algorithm~\ref{alg:GivensFwdRRUnitary}.

\begin{algorithm}[H]
  \caption{Parallel Forward $\Uv \in \U(n)$}
  \label{alg:GivensFwdRRUnitary}
  \begin{algorithmic}[0] 
    \State \textbf{Input:}
    \State \hspace*{\algorithmicindent} Block sequence $B = ( b_1,
    \ldots, b_{n-1})$ such that $E$ is round-robin
    \State \hspace*{\algorithmicindent} Parametrizing angles $\thetav = (\theta_{e_1}, \ldots, \theta_{e_\Ncal})$
    \State \hspace*{\algorithmicindent} Parametrizing phase angles $\phiv = (\phi_{e_1}, \ldots, \phi_{e_\Ncal})$
    \State \textbf{Output:}
    \State \hspace*{\algorithmicindent} $\Uv(\thetav)$
    \State $\Uv \gets \Iv_n$
    \For{$b \in \textrm{reversed}(B)$}
      \For{$e \in b$} \Comment{This block is fully parallel}
        \State $(i,j) \gets e$
        \State $\rv_i \gets e^{\imag\phi_{ij}}\cos\theta_{ij}\Uv_{i:} -
        \sin\theta_{ij}\Uv_{j:}$
        \State $\rv_j \gets e^{\imag\phi_{ij}} \sin\theta_{ij}\Uv_{i:} +
        \cos\theta_{ij}\Uv_{j:}$
        \State $\Uv_{i:} \gets \rv_i$
        \State $\Uv_{j:} \gets \rv_j$
      \EndFor
      \State Synchronize parallel operations
    \EndFor
  \end{algorithmic}
\end{algorithm}

Extending the backpropagation computation takes a bit more work.
The matrix 
\[
  \frac{\partial\Gv^e}{\partial\theta_e}
\]
consists of 4 nonzero elements
\begin{align}
  \Bigg[ \frac{\partial\Gv^e}{\partial\theta_e} \Bigg]_{ii} = & -e^{\imag\phi_{e}}\sin\theta_e & \Bigg[ \frac{\partial\Gv^e}{\partial\theta_e} \Bigg]_{jj} = & -\sin \theta_e \nonumber  \\
  \Bigg[ \frac{\partial\Gv^e}{\partial\theta_e} \Bigg]_{ij} = & -\cos\theta_e  &   \Bigg[ \frac{\partial\Gv^e}{\partial\theta_e} \Bigg]_{ji} = & e^{\imag\phi_{e}}\cos\theta_e
\end{align}
while 
\[
\frac{\partial\Gv^e}{\partial\phi_e}
\]
possesses two nonzero elements
\begin{align}
  \Bigg[ \frac{\partial\Gv^e}{\partial\phi_e} \Bigg]_{ii} = & \imag e^{\imag\phi_{e}}\cos\theta_e &   \Bigg[ \frac{\partial\Gv^e}{\partial\phi_e} \Bigg]_{ji} = & \imag e^{\imag\phi_{e}}\sin \theta_e 
\end{align}
For block $b_k$ and $\tilde{e} \in b_k$, commutativity implies that
\begin{align}
 \frac{\partial \Gv^{b_k}}{\partial\theta_{\tilde{e}}} = \frac{\partial\Gv^{\tilde{e}}}{\partial\theta_{\tilde{e}}} \Gv^{\tilde{e}}{}^\dagger \Gv^{b_k} \nonumber \\
 \frac{\partial \Gv^{b_k}}{\partial\phi_{\tilde{e}}} = \frac{\partial\Gv^{\tilde{e}}}{\partial\phi_{\tilde{e}}} \Gv^{\tilde{e}}{}^\dagger \Gv^{b_k}
\end{align}
where $\dagger$ denotes the adjoint (conjugate/transpose) operation. Define
\begin{align}
  \Qv_e \triangleq \frac{\partial\Gv^e}{\partial\theta_e} \Gv^e{}^\dagger \nonumber \\
  \Pv_e \triangleq \frac{\partial\Gv^e}{\partial\phi_e} \Gv^e{}^\dagger
\end{align}
Just as for the real case, the entries of $\Qv_e$ are only nonzero in two locations, namely
\begin{align}
  \Big[ \Qv_e\Big]_{ij} & = -1 \nonumber \\
  \Big[ \Qv_e\Big]_{ji} & = 1
\end{align}
while $\Pv_e$ has four (purely imaginary) nonzero entries:
\begin{align}
  \Big[ \Pv_e\Big]_{ii} & =  \imag \cos^2\theta_e \nonumber \\
  \Big[ \Pv_e\Big]_{jj} & =  \imag  \sin^2\theta_e \nonumber \\
  \Big[ \Pv_e\Big]_{ij} & =  \Big[ \Pv_e\Big]_{ji}  = \imag   \sin\theta_e \cos\theta_e 
\end{align}

At step $k$, again let $\Uv^\fwd = \Uv^{1:k-1}$ and $\Uv^\bck = \Uv^{k:n-1}$, with
\[
  \Uv^\fwd = [\uv_1 \uv_2 \ldots \uv_n]
\]
and
\[
  \Uv^\bck = \begin{bmatrix}
    \vv^\dagger_1 \\
    \vv^\dagger_2 \\
    \vdots \\
    \vv^\dagger_n
\end{bmatrix}
\]
Turning back to obtaining
\begin{align}
  \frac{\partial\Uv}{\partial\theta_e} = & \Uv^\fwd \Qv_e \Uv^\bck \nonumber \\
  \frac{\partial\Uv}{\partial\phi_e} = & \Uv^\fwd \Pv_e \Uv^\bck
\end{align}
the properties of $\Qv_e$ and $\Pv_e$ imply that for $e \in b_k$
\begin{align}
  \frac{\partial\Uv}{\partial\theta_e} = & \uv_j\vv_i^\dagger - \uv_i\vv_j^\dagger \nonumber \\
  \frac{\partial\Uv}{\partial\phi_e} = & \imag\cos \theta_e\big( \cos\theta_e \uv_i + \sin\theta_e \uv_j \big) \vv_i^\dagger + \imag\sin \theta_e\big( \cos\theta_e \uv_i + \sin\theta_e \uv_j \big) \vv_j^\dagger \nonumber \\
  = & \imag \big( \cos\theta_e \uv_i + \sin\theta_e \uv_j \big) \big( \cos\theta_e \vv_i + \sin\theta_e \vv_j \big)^\dagger
\end{align}
Thus, just as for the real case the derivative matrices with respect
to $\{\theta_e\}$ are of rank 2, while those with respect to the phase
angles $\{\phi_e\}$ have unity rank; within a block, disjoint pairs of
$(\uv_i, \uv_j)$ and $(\vv_i,\vv_j)$ are involved in the computations
of both derivative types.

Completion of the block-parallel gradient computation can then follow
an analogous path to that described in
Section~\ref{sec:JVP:ComputingGrad}, albeit more tediously.

\end{appendices}

\end{document}